\documentclass[10pt]{article}

\usepackage{amsmath,amssymb,amsthm,bm}
\usepackage{booktabs}
\usepackage{graphicx}
\usepackage{microtype}
\usepackage{xcolor}
\usepackage{array}
\usepackage{enumitem}
\usepackage{seqsplit}
\usepackage[margin=1in]{geometry}
\usepackage[authoryear,round]{natbib}
\usepackage[colorlinks=true,linkcolor=black,citecolor=blue!50!black,urlcolor=blue!50!black]{hyperref}
\usepackage[nameinlink,noabbrev]{cleveref}
\usepackage{url}
\usepackage[section]{placeins}
\usepackage{rotating}
\providecommand{\doi}[1]{\href{https://doi.org/#1}{\nolinkurl{doi:#1}}}
\newcommand{\breakhash}[1]{\texttt{\expandafter\seqsplit\expandafter{#1}}}
\newcommand{\srank}{\operatorname{srank}}
\newcommand{\Q}{Q}
\newcommand{\Xt}{\tilde X}

\graphicspath{{figures/}}
\setlist{nosep,leftmargin=*}

\providecommand{\FreezeCommit}{e2be93a7052c2998f52543db78f0457815d82569}

\providecommand{\PreregSha}{919b08148f80196daa97a51fc8fbb5cccdea729a204596cdf29da03ff8744d61}
\providecommand{\CalibrationSha}{0c735b79cff13134a45c8ead5b130a4431aa8ee605b023dca1e7a49f4d14175b}
\providecommand{\AlphaMin}{0.02}
\providecommand{\Budget}{4,000}
\providecommand{\Lam}{0.1}
\providecommand{\RampStart}{1,500}
\providecommand{\RampEnd}{3,000}

\providecommand{\SrankRecTwelve}{3.95}

\providecommand{\SrankRecTwentyFour}{3.86}

\providecommand{\NumSettled}{8}
\providecommand{\NumTransient}{2}

\providecommand{\LossGapMin}{1.73}
\providecommand{\LossGapMax}{1.94}
\providecommand{\NPenOne}{8}
\providecommand{\RecPenOne}{0}

\providecommand{\NHingeOne}{8}
\providecommand{\RecHingeOne}{0}

\providecommand{\NPenFour}{8}
\providecommand{\RecPenFour}{0}

\providecommand{\NHingeFour}{8}
\providecommand{\RecHingeFour}{0}

\providecommand{\NCosOne}{8}
\providecommand{\RecCosOne}{0}

\providecommand{\NRampup}{8}
\providecommand{\RecRampup}{8}

\providecommand{\NoneNotRecoveredTwelve}{3/3}
\providecommand{\NoneNotRecoveredTwentyFour}{5/5}
\providecommand{\RhoPenOne}{0.0032}

\providecommand{\RhoHingeOne}{0.0887}

\providecommand{\RhoHingeFour}{0.3071}

\providecommand{\RhoCosOne}{0.0618}

\providecommand{\RhoRatioHingePen}{28}

\providecommand{\AlphaRecTwelveMin}{0.25}
\providecommand{\AlphaRecTwelveMax}{0.97}
\providecommand{\AlphaRecTwentyFourMin}{1.00}

\providecommand{\MarkerOrderOk}{0/10}
\providecommand{\LossNotLater}{10/10}
\providecommand{\MarkerGapMin}{2,990}
\providecommand{\MarkerGapMax}{3,030}
\providecommand{\MarkerUMin}{2,980}
\providecommand{\MarkerUMax}{3,040}
\providecommand{\MarkerDlossMin}{2,960}
\providecommand{\MarkerDlossMax}{3,000}
\providecommand{\MatchedWithinTen}{31/40}

\providecommand{\CalCollapseAtThree}{2/6}
\providecommand{\CalCollapseAtTwo}{6/6}
\providecommand{\CalCollapseAtZero}{6/6}
\providecommand{\CalNoneForks}{12/12}
\providecommand{\CalFunctionForksN}{23}

\providecommand{\GpuHours}{54}
\providecommand{\HealthyN}{26}
\providecommand{\HealthyDipN}{24}
\providecommand{\HealthyDipMax}{160}
\providecommand{\HealthyRecMax}{550}
\providecommand{\ParamsTwelve}{9}
\providecommand{\ParamsTwentyFour}{11}

\providecommand{\HealthyAtCollapsedLossTwelve}{24}
\providecommand{\HealthyAtCollapsedGradTwelve}{265}

\providecommand{\HealthyAtCollapsedLossTwentyFour}{59}

\providecommand{\AuditPenForceHigh}{$4.9\cdot10^{-4}$}
\providecommand{\AuditPenForceLow}{$3.5\cdot10^{-8}$}
\providecommand{\AuditHingeForce}{0.022}
\providecommand{\AuditDeltaHigh}{1}
\providecommand{\AuditDeltaLow}{0.0001}

\providecommand{\AuditRhoPen}{0.0013}
\providecommand{\AuditRhoHinge}{0.074}

\providecommand{\UnigramEntropy}{7.64}
\providecommand{\CollapsedLossMin}{7.55}
\providecommand{\CollapsedLossMax}{7.63}
\providecommand{\HingeSrankFirstMin}{8}
\providecommand{\HingeSrankFirstMax}{9}
\providecommand{\HingeSrankMidMin}{1.6}
\providecommand{\HingeSrankMidMax}{2.7}
\providecommand{\RampUpStepsToThree}{430}

\IfFileExists{generated/posthoc_macros.tex}{
\providecommand{\HoldOneN}{8}
\providecommand{\HoldOneRec}{2}

\providecommand{\HoldThreeN}{8}
\providecommand{\HoldThreeRec}{7}
\providecommand{\HoldThreeTmin}{34}
\providecommand{\HoldThreeTmax}{121}
\providecommand{\HoldThreeRecTwelve}{3}
\providecommand{\HoldThreeNTwelve}{3}
\providecommand{\HoldThreeRecTwentyFour}{4}
\providecommand{\HoldThreeNTwentyFour}{5}
\providecommand{\HoldSixN}{8}
\providecommand{\HoldSixRec}{8}

\providecommand{\AgeFiveTmin}{130}
\providecommand{\AgeFiveTmax}{170}
\providecommand{\AgeFiveRecTwelve}{2}
\providecommand{\AgeFiveNTwelve}{3}
\providecommand{\AgeFiveRecTwentyFour}{0}
\providecommand{\AgeFiveNTwentyFour}{5}

\providecommand{\AgeFifteenTmin}{240}
\providecommand{\AgeFifteenTmax}{440}
\providecommand{\AgeFifteenRecTwelve}{3}
\providecommand{\AgeFifteenNTwelve}{3}
\providecommand{\AgeFifteenRecTwentyFour}{0}
\providecommand{\AgeFifteenNTwentyFour}{5}

\providecommand{\AgeThirtyTmin}{3770}

\providecommand{\AgeThirtyRecTwelve}{1}
\providecommand{\AgeThirtyNTwelve}{3}
\providecommand{\AgeThirtyRecTwentyFour}{0}
\providecommand{\AgeThirtyNTwentyFour}{5}

\providecommand{\NoclipN}{32}
\providecommand{\NoclipRec}{0}
\providecommand{\EntActMin}{0.94}
\providecommand{\EntActMax}{0.99}
\providecommand{\EntHealthyMin}{0.65}
\providecommand{\EntHealthyMax}{0.69}
\providecommand{\EntLateMin}{0.94}
\providecommand{\EntLateMax}{0.99}
\providecommand{\RepNSettled}{10}

\providecommand{\RepAlphaRecTwelveMin}{0.95}
\providecommand{\RepAlphaRecTwelveMax}{1.00}

\providecommand{\RepAlphaRecTwentyFourMin}{1.00}

\providecommand{\RepRhoPen}{0.0037}
\providecommand{\RepRhoHinge}{0.0782}
\providecommand{\HystEventsN}{18}
\providecommand{\HystMinRatio}{12}
\providecommand{\RepRampupRec}{10}
\providecommand{\RepRampupN}{10}

\providecommand{\SkipRecN}{42}
\providecommand{\SkipRecGapMin}{0.21}
\providecommand{\SkipRecGapMax}{0.56}
\providecommand{\SkipRecGapMed}{0.44}

\providecommand{\AlphaDetMin}{0.02}
\providecommand{\AlphaDetMax}{0.03}

}{}
\providecommand{\PathN}{18}

\providecommand{\PathMOneTaskOk}{17}

\providecommand{\PathAttnLastMedTwelve}{$8\cdot10^{-3}$}

\providecommand{\PathAttnFirstMedTwelve}{$10^{-8}$}

\providecommand{\PathAttnHealthyMedTwelve}{$7\cdot10^{-2}$}
\providecommand{\PathHingeFirstMedTwelve}{0.34}
\providecommand{\PathHingeLastMedTwelve}{$6\cdot10^{-4}$}

\providecommand{\PathQkHealthyMedTwelve}{$5\cdot10^{-2}$}

\providecommand{\PathAttnLastMedTwentyFour}{$5\cdot10^{-3}$}

\providecommand{\PathAttnFirstMedTwentyFour}{$2\cdot10^{-11}$}

\providecommand{\PathQkHealthyMedTwentyFour}{$3\cdot10^{-2}$}

\providecommand{\PathQkTinyN}{17}

\providecommand{\PathHealthyQkMinOk}{12}

\providecommand{\PathSwitchThreeStepZeroOk}{1}

\providecommand{\PathSwitchThreeStepZeroAttnOk}{9}

\providecommand{\PathSwitchThreeAfterFiftyAboveRec}{9}
\providecommand{\PathSwitchThreeStepZeroAboveRec}{0}

\providecommand{\PathSwitchOneStepZeroOk}{15}

\providecommand{\PathSwitchOneStepZeroAttnOk}{18}

\providecommand{\PathSwitchOneRiseMed}{7}
\providecommand{\PathSwitchOneVsHealthyMed}{1.2}

\providecommand{\PathSwitchOneAfterFiftyAboveRec}{8}
\providecommand{\PathSwitchOneStepZeroAboveRec}{4}

\providecommand{\PathForkCheckMedPct}{2}

\IfFileExists{generated/excursion_macros.tex}{

}{}
\IfFileExists{generated/excursion_test_macros.tex}{
\providecommand{\ExcTestN}{8}

\providecommand{\ExcTestQkSixteenMin}{$1.9\cdot10^{-4}$}
\providecommand{\ExcTestQkSixteenMax}{$7.1\cdot10^{-4}$}
\providecommand{\ExcTestRecSixteen}{8}

\providecommand{\ExcTestTOneSixteen}{8}
\providecommand{\ExcTestTOneSixteenMet}{met}
\providecommand{\ExcTestTrecSixteenMin}{1110}
\providecommand{\ExcTestTrecSixteenMax}{3540}

\providecommand{\ExcTestQkThirtyTwoMin}{$1.7\cdot10^{-5}$}
\providecommand{\ExcTestQkThirtyTwoMax}{$1.5\cdot10^{-4}$}

\providecommand{\ExcTestTOneThirtyTwo}{3}
\providecommand{\ExcTestTOneThirtyTwoMet}{missed}

\providecommand{\ExcTestTZero}{8}
\providecommand{\ExcTestTZeroMet}{met}
\providecommand{\ExcTestTOneB}{16}
\providecommand{\ExcTestTOneBN}{16}
\providecommand{\ExcTestTOneBMet}{met}
\providecommand{\ExcTestTTwoLoss}{14}
\providecommand{\ExcTestTTwoGrad}{12}

\providecommand{\ExcTestTTwoMet}{missed}
\providecommand{\ExcTestTThreeN}{8}
\providecommand{\ExcTestTThreeRho}{$-0.10$}
\providecommand{\ExcTestTThreeMet}{missed}
\providecommand{\ExcTestTThreeLossRho}{$+0.79$}
\providecommand{\ExcTestTThreeGradRho}{$+0.55$}
\providecommand{\ExcTestTThreeRMet}{missed}
\providecommand{\ExcTestTThreePQk}{$-0.82$}
\providecommand{\ExcTestTThreePLoss}{$+0.89$}

\providecommand{\ExcTestTThreePMult}{$+0.93$}
\providecommand{\ExcTestEndThirtyTwoMin}{2.7}
\providecommand{\ExcTestEndThirtyTwoMax}{3.4}
\providecommand{\ExcTestTFourRec}{8}

\providecommand{\ExcTestTFourMet}{met}
\providecommand{\ExcTestTFourGapMin}{0.86}
\providecommand{\ExcTestTFourGapMax}{1.04}
}{}
\IfFileExists{generated/prediction_macros.tex}{
\providecommand{\PredThreshold}{$3.1\cdot10^{-5}$}
\providecommand{\PredMainAcc}{8/8}
\providecommand{\PredDevRho}{$-0.79$}
\providecommand{\PredN}{10}
\providecommand{\PredPOneCorrect}{8}

\providecommand{\PredPOneObsRec}{5}
\providecommand{\PredPOneObsRecTwelve}{3}
\providecommand{\PredNTwelve}{5}
\providecommand{\PredPOneObsRecTwentyFour}{2}
\providecommand{\PredNTwentyFour}{5}
\providecommand{\PredPTwoN}{9}
\providecommand{\PredPTwoRho}{$-0.67$}

\providecommand{\PredRivalHingeAcc}{5}
\providecommand{\PredRivalHingeRho}{$+0.45$}
\providecommand{\PredRivalLossGapAcc}{8}
\providecommand{\PredRivalLossGapRho}{$-0.68$}
\providecommand{\PredRivalGradNormAcc}{4}
\providecommand{\PredRivalGradNormRho}{$-0.08$}
\providecommand{\PredRivalDepthAcc}{6}

\providecommand{\RepAgeFiveRecTwelve}{3}

\providecommand{\RepAgeFiveTmin}{110}
\providecommand{\RepAgeFiveTmax}{180}
\providecommand{\RepAgeFifteenRecTwelve}{2}

\providecommand{\RepAgeFifteenTmin}{260}
\providecommand{\RepAgeFifteenTmax}{300}
\providecommand{\RepAgeThirtyRecTwelve}{0}

\providecommand{\RepHoldThreeRecTwelve}{5}
\providecommand{\RepHoldThreeRecTwentyFour}{4}
}{}

\title{Force without transmission:\\
a depth-induced rank collapse\\ that no loss on the representation reopens}

\author{%
  Martin Hofmann$^{1}$ \qquad Patrick M\"ader$^{1,2,3}$\\[0.6em]
  \small $^{1}$ Data-intensive Systems and Visualization Group (dAI.SY),
  Technische Universit\"at Ilmenau,\\
  \small Max-Planck-Ring 14, 98693 Ilmenau, Thuringia, Germany\\
  \small $^{2}$ German Centre for Integrative Biodiversity Research (iDiv)
  Halle--Jena--Leipzig,\\
  \small Deutscher Platz 5e, 04103 Leipzig, Saxony, Germany\\
  \small $^{3}$ Faculty of Biological Sciences, Friedrich Schiller University,\\
  \small F\"urstengraben 1, 07745 Jena, Thuringia, Germany\\[0.6em]
  \small \href{mailto:martin.hofmann@tu-ilmenau.de}{\texttt{martin.hofmann@tu-ilmenau.de}}
  \qquad
  \href{mailto:patrick.maeder@tu-ilmenau.de}{\texttt{patrick.maeder@tu-ilmenau.de}}\\
  \small ORCID:
  \href{https://orcid.org/0000-0002-4440-3317}{0000-0002-4440-3317}
  \qquad
  \href{https://orcid.org/0000-0001-6871-2707}{0000-0001-6871-2707}
}

\date{October 2026}

\begin{document}
\maketitle

\begin{abstract}
Training can drive a transformer into a rank collapse: all token representations point in one direction, and
learning stops.  In a related collapse of attention, a loss term with a bounded corrective
force repairs the network during the run.  We ask whether such a term repairs rank collapse.
We collapse small transformers by weakening their skip connection and treat copies of the collapsed network.  No added loss term repaired the collapse, although the stronger kind pushed with
about a tenth of the task gradient.  The reason was the path, not the strength.  The task gradient no longer
reached the query and key weights, which decide where attention looks, and the added term's gradient faded
before the blocks where the collapse forms.  Restoring the skip connection, which changes no weight, reopened
this path at once.  The rank then recovered, but only far above the scale of collapse.  After a burst of high learning rate the path stayed open and the rank recovered
untreated.  Registered predictions from the path ranked recovery times but did not transfer to this cause.  In every case the loss stayed above that of a healthy network after the rank recovered.  Whether a
collapsed network can be repaired depends on whether the gradient still reaches the weights that must change,
not on how strongly a loss term pushes.
\end{abstract}

\section{Introduction}
\label{sec:intro}

A transformer keeps tokens apart by passing each token's own representation forward through a skip connection
and adding what attention and the feed-forward layer compute.  If the skip connection is weakened enough,
attention averages the tokens again and again across the blocks, and their representations converge onto a
single direction.  This is depth-induced rank collapse \citep{dong2021attention}.  The network can no longer
tell tokens apart, and training does not lead it out: in that state the gradient that would sharpen attention
vanishes \citep{noci2022signal}.  The known defences act before the fact, by changing the architecture or the
optimizer \citep{zhai2023stabilizing,henry2020query,noci2022signal,ren2026msign}.

Once a run has collapsed, the architecture is fixed and the cheapest lever left is an extra term in the loss.
Whether such a term can repair the collapse is not obvious.  For the related collapse of attention entropy the
answer depends on what the term measures.  A term built on the quantity that has collapsed loses its push
exactly there, because its gradient fades to zero; a term built on a spectral quantity keeps a push of bounded
size \citep{zhai2023stabilizing,bardes2022vicreg,mei2020escaping,rusch2022graphcon}.  We call the size of the
gradient a term produces at the weights its \emph{force}, and the two kinds a \emph{vanishing} and a
\emph{bounded} force.  In attention-entropy collapse a term of the bounded kind does
repair an established collapse from saved checkpoints \citep{hofmann2026repairing}.  This suggests a simple
rule: choose a term whose force does not vanish.  But a force exerted on the loss acts on the weights only
through the backward pass, the same chain of blocks that has collapsed.  A large force at the loss is
therefore not the same as a force that arrives.  We call the part of a gradient that reaches the weights of a
given block its \emph{transmission}, and the route it takes through the stack its \emph{path}.

We test which of the two decides.  We collapse small transformers by weakening their skip connection, save the
collapsed network, and continue copies of it under different treatments: two loss terms that differ only in
whether their force vanishes, an untreated copy, and copies in which the skip connection is restored.  We
measure how large each force is and, block by block, how much of it reaches the attention weights.  We then
test, with predictions registered before the runs, whether this measurement of the path tells in advance
which networks will recover, both on new networks and under a second cause of collapse, a short burst of high
learning rate.  A second question concerns time: the theory of rank collapse is stated at initialisation
\citep{dong2021attention,noci2022signal,giorlandino2026two}, and we ask whether it matters how long a network
has been collapsed before it is treated \citep[compare][]{ersoy2026noise}.

The question matters beyond this failure.  Auxiliary losses, regularisers and penalties are routinely added to
steer a network, and they are usually judged by their size at the loss.  Every such term reaches the weights
through the backward pass, so whether it can act depends on the state of that pass.  A collapse is a case in
which the state of the pass can be changed by one runtime setting and measured directly.  The paper makes four
contributions:

\begin{enumerate}
\item two loss terms on the rank of the token representations that differ only in whether their force
  vanishes at the collapse, tested as repairs of a settled rank collapse;
\item a block-by-block measurement of where the task gradient and the added gradient arrive inside the
  collapsed network, before and after the skip connection is restored;
\item a test of how the time spent collapsed changes what a restoration of the skip connection achieves;
\item two registered tests of the path measurement as a predictor of recovery.
\end{enumerate}

\section{Two kinds of corrective force}
\label{sec:classes}

To compare a vanishing with a bounded force, we need two loss terms that measure the same thing and differ
only in how their gradient behaves at the collapse.  The thing measured is how many directions the token
representations of a block span.  For a block's token matrix $X\in\mathbb{R}^{T\times d}$, centred over tokens
to $\Xt$ with singular values $\sigma_1\ge\dots\ge\sigma_d$, the stable rank
$\srank(\Xt)=\|\Xt\|_F^2/\sigma_1^2$, in short the \emph{rank}, is 1 when all tokens point in one direction and
grows with the number of directions they span.  With $E=\Xt-\sigma_1u_1v_1^\top$ the part outside the top
direction and $\Q=\sqrt{\srank-1}=\|E\|_F/\sigma_1$ its relative size, the two terms are
\begin{align}
  R_{\mathrm{pen}}(\Xt)   &= 1-\frac{\Q^2}{d-1}, &
  R_{\mathrm{hinge}}(\Xt) &= \max(0,\,s_0-\Q),
\end{align}
where $s_0$ is the healthy value of $\Q$.  The \emph{penalty} is the rank itself, rescaled to 1 at the
collapse.  Its gradient is proportional to $\Q$ and vanishes as the rank approaches one.  The \emph{hinge}
acts only below the healthy value, and its gradient keeps a constant size $1/\sigma_1$ in the direction that
would spread the tokens, because $\nabla_{\Xt}\Q=E/(\|E\|_F\,\sigma_1)-(\|E\|_F/\sigma_1^2)\,u_1v_1^\top$.
Both terms act on every block, averaged over blocks, and each enters the loss with a weight chosen so that it
adds the same amount to the loss when it is switched on.  A third term, the mean squared cosine between token
pairs, is a common choice in the literature and serves as a comparison.

\section{How the collapse is produced and treated}
\label{sec:setup}

\begin{table}[t]
  \centering
  \caption{Treatments started from each saved collapsed network.  $s_0=\sqrt{\srank_{\mathrm{healthy}}-1}$.}
  \label{tab:arms}
  \begin{tabular}{llll}
  \toprule
  Treatment & Added loss & Force class & Added loss when switched on \\
  \midrule
  penalty $1\times$ / $4\times$ & $\beta\,(1-\Q^2/(d-1))$ & vanishing & $\Lambda$ / $4\Lambda$ \\
  hinge $1\times$ / $4\times$ & $\beta\max(0,s_0-\Q)$ & bounded & $\Lambda$ / $4\Lambda$ \\
  cosine $1\times$ & $\beta\,U$ & neither & $\Lambda$ \\
  untreated & -- & control & -- \\
  reversed ramp & -- & driver reversed & -- \\
  \bottomrule
  \end{tabular}
\end{table}

We need a collapse that is reproducible, settled and harmful, and a way to apply every treatment to the same
collapse.  We train small language models: pre-LN decoder transformers of width 128 with four heads, context
256 and the GPT-2 vocabulary, at depths 12 and 24 (about \ParamsTwelve{} and \ParamsTwentyFour\,M parameters),
with AdamW at a learning rate of $10^{-3}$ and gradient clipping at norm 1, on a fixed stream of FineWeb-Edu
batches \citep{penedo2024fineweb}.  Query--key normalisation \citep{henry2020query} prevents the other kind of
collapse, attention entropy collapse; we checked that it did not occur.

Each block computes $x\leftarrow\alpha x+\mathrm{Attn}(\mathrm{LN}(x))$ and
$x\leftarrow\alpha x+\mathrm{MLP}(\mathrm{LN}(x))$, so $\alpha$ scales the skip connection.  It is 1 for the
first \RampStart{} steps and then ramps down to $\alpha_{\min}=\AlphaMin$ by step \RampEnd{}.  We call this
ramp the \emph{driver} of the collapse.  The rank of the last block is measured every ten steps on a fixed
probe batch.  A network counts as collapsed below a rank of 1.5 and is saved once the collapse has settled.
It counts as recovered when the 50-step median of its rank reaches half the healthy value; recovery in this
paper always means recovery of the rank, and the loss is reported separately.  Each treatment continues the
saved network for \Budget{} steps.  The details of calibration and detection are in \cref{app:design}.

The treatments (\cref{tab:arms}) are the penalty and the hinge at one and four times a matched added loss of
$\Lambda=\Lam$ nats, the cosine term, an untreated copy, and the \emph{reversed ramp}, in which $\alpha$
returns to 1 at the rate it came down.  Setting $\alpha$ back is a runtime action: it changes no weight, only
how much of each block's input passes through.  Two depths and five seeds give ten collapse events; a
replication with five new seeds per depth followed under the same rules.  The treatments and their evaluation
were fixed before the main runs.  After them we added branches that set $\alpha$ back to a fixed value at once
or after a delay, the block-by-block gradient measurements, and two tests whose predictions were committed to
the repository before their runs; all of these are marked where they appear.

\section{The collapse is harmful and does not heal}
\label{sec:collapse}

Before asking what repairs the collapse, we need to know that it is damage and that it stays.  When it was
saved, the collapsed network's loss lay \LossGapMin--\LossGapMax{} nats above that of a healthy network
trained on the same data, at \CollapsedLossMin--\CollapsedLossMax{} nats, close to the \UnigramEntropy{} nats of
a model that only knows how often each token occurs.  Left untreated for \Budget{} steps, the rank stayed
below the recovery threshold in \NoneNotRecoveredTwelve{} collapse events at depth 12 and
\NoneNotRecoveredTwentyFour{} at depth 24, and in all \RepNSettled{} events of the replication.  Attention in
the collapsed state was close to uniform: every token attended to all tokens almost equally (normalised
entropy \EntActMin--\EntActMax, against \EntHealthyMin--\EntHealthyMax{} in healthy networks).  The collapse
was a stable, harmful state that training alone did not leave.

\section{No added loss repairs it, although its force is large}
\label{sec:repair}

\begin{figure}[t]
  \centering
  \includegraphics[width=\linewidth]{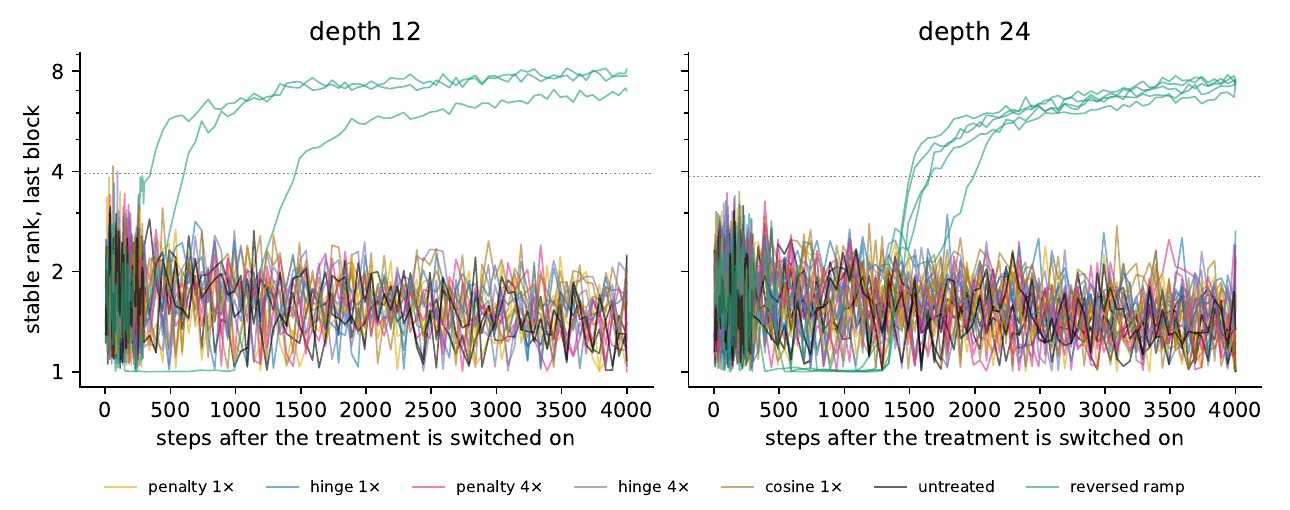}
  \caption{Stable rank of the last block after the treatment is switched on, all treatments and collapse
  events of the main study (50-step median, as in the recovery rule; dotted line: recovery threshold).  Only
  the reversed ramp recovers.}
  \label{fig:arms}
\end{figure}

The first question is whether either kind of loss term repairs the collapse.  All loss-based branches ran with
the skip scale held at $\alpha_{\min}$, so each term worked against the cause of the collapse.  None recovered
the rank: penalty \RecPenOne/\NPenOne, hinge \RecHingeOne/\NHingeOne, penalty $4\times$ \RecPenFour/\NPenFour,
hinge $4\times$ \RecHingeFour/\NHingeFour{} and cosine \RecCosOne/\NCosOne{} (\cref{fig:arms}), and none of the
loss-based branches of the replication.  Exempting the added gradient from clipping, at up to ten times the
strength, changed nothing (\NoclipRec{} of \NoclipN{} branches recovered).  The loss terms also did not merely
hasten a recovery that would have come anyway, because the untreated copies did not recover either.

A failed repair could mean that the force was too weak.  We therefore measured it at the parameters, as the
ratio $\rho$ of the added gradient to the task gradient.  The penalty's force faded as designed, and the
hinge's stayed: the median per collapse event was \RhoPenOne{} for the penalty and \RhoHingeOne{} for the
hinge, about \RhoRatioHingePen{} times more, and \RhoHingeFour{} for the hinge at four times the strength
(\cref{app:force}; replication \RepRhoPen{} and \RepRhoHinge).  The distinction between the two kinds of force
thus held for rank collapse, and the bounded force was large.  It still did not repair the collapse.

\section{The force does not reach the weights that must change}
\label{sec:where}

\begin{figure}[t]
  \centering
  \includegraphics[width=\linewidth]{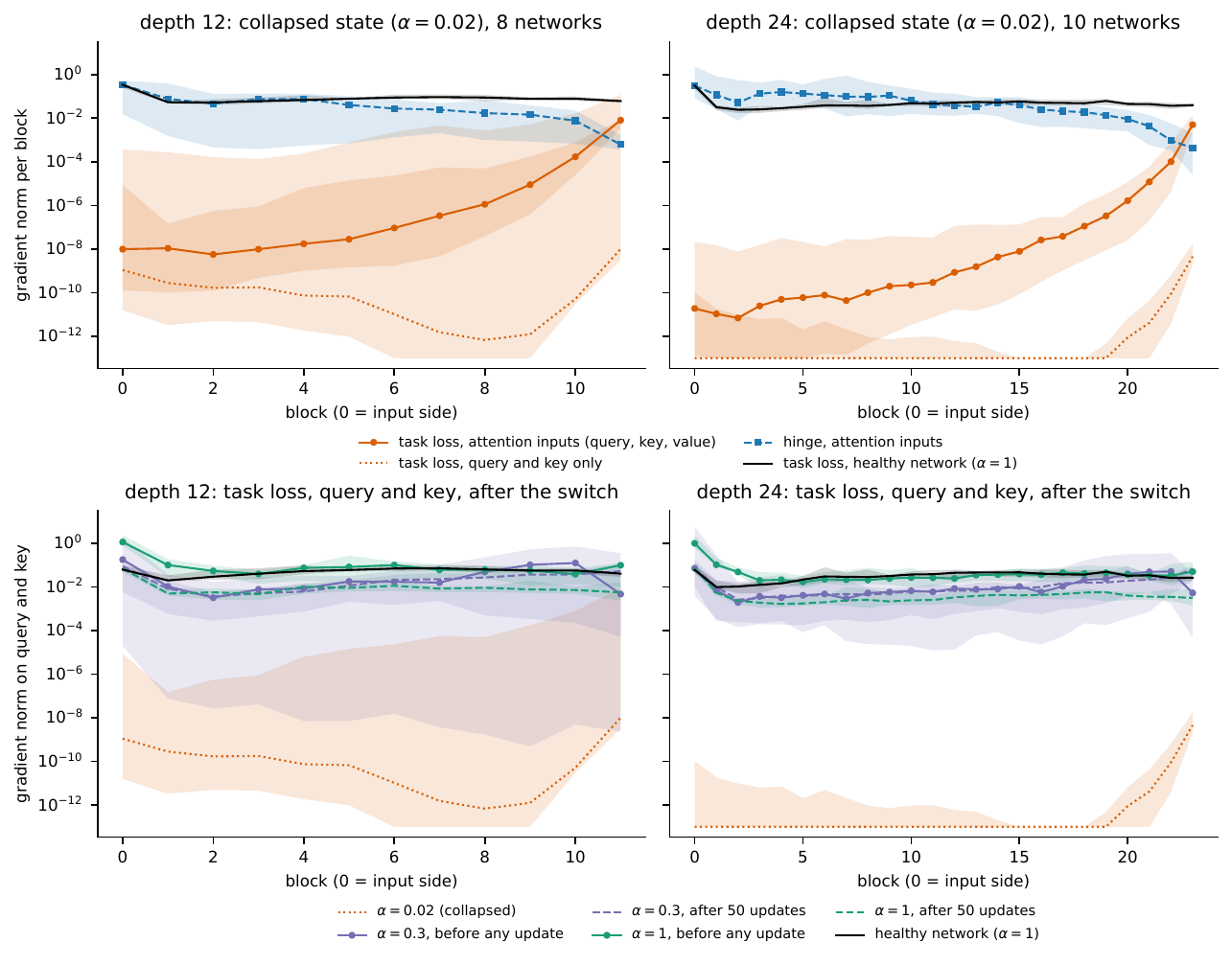}
  \caption{Gradient path in the saved collapsed networks of the main study and the replication (lines:
  median over networks; bands: range).  Top: per-block gradient norm at the collapsed state for the task loss
  on the attention input weights (solid) and on the query and key part alone (dotted), for the per-block
  hinge at matched added loss (dashed), and for the task loss in the healthy network of the same seed
  (black).  Bottom: task gradient on the query and key weights after the skip scale is set to 0.3 or 1, at
  the first step before any update (solid) and after 50 updates of the branch (dashed).  Values below
  $10^{-13}$ are numerically zero in single precision and are drawn at the lower edge.}
  \label{fig:path}
\end{figure}

If the force is large but has no effect, it may not arrive where it is needed.  For tokens to become distinct
again, attention must stop averaging them, so the weights that decide where attention looks, the query and
key projections, have to change.  We measured, block by block, how much of the task gradient and of the
hinge's gradient reached the attention weights of each block, on all \PathN{} saved collapsed networks of the
main study and the replication, without any training (\cref{fig:path}, top).  The healthy reference is the
network of the same seed trained without the driver.

The task gradient arrived weakly at the blocks nearest the output and hardly at all at those nearest the
input.  On a block's attention weights (query, key and value) it was \PathAttnLastMedTwelve{} in the last block
and \PathAttnFirstMedTwelve{} in the first at depth 12, against about \PathAttnHealthyMedTwelve{} per block in
a healthy network; at depth 24 it ran from \PathAttnLastMedTwentyFour{} in the last block to
\PathAttnFirstMedTwentyFour{} in the first, and it fell by more than three orders of magnitude from the last block to the first in
\PathMOneTaskOk{} of \PathN{} networks.  Almost all of it sat on the value projection.  The query and key
weights received less than $10^{-6}$ in every block of \PathQkTinyN{} of \PathN{} networks, against
\PathQkHealthyMedTwentyFour{} to \PathQkHealthyMedTwelve{} per block in healthy ones.  This is the result that
\citet{noci2022signal} derive at initialisation, here in trained networks: when all tokens look alike,
changing where a token attends changes nothing, so no gradient asks for it.

The hinge's gradient had the opposite profile.  It was \PathHingeFirstMedTwelve{} in the first block at depth
12, more than the healthy task gradient, and \PathHingeLastMedTwelve{} in the last, because each block's term
can only act through the diversity its own input still has.  The blocks that complete the collapse, near the
output, therefore received almost nothing from either side.  The hinge branches show the consequence: at the
end of the budget the rank was \HingeSrankFirstMin--\HingeSrankFirstMax{} in the first block but
\HingeSrankMidMin--\HingeSrankMidMax{} in the middle one, so the diversity the hinge restored near the input was
lost again on the way to the output.

\section{Restoring the path recovers the rank, with hysteresis}
\label{sec:runtime}

\begin{figure}[t]
  \centering
  \includegraphics[width=\linewidth]{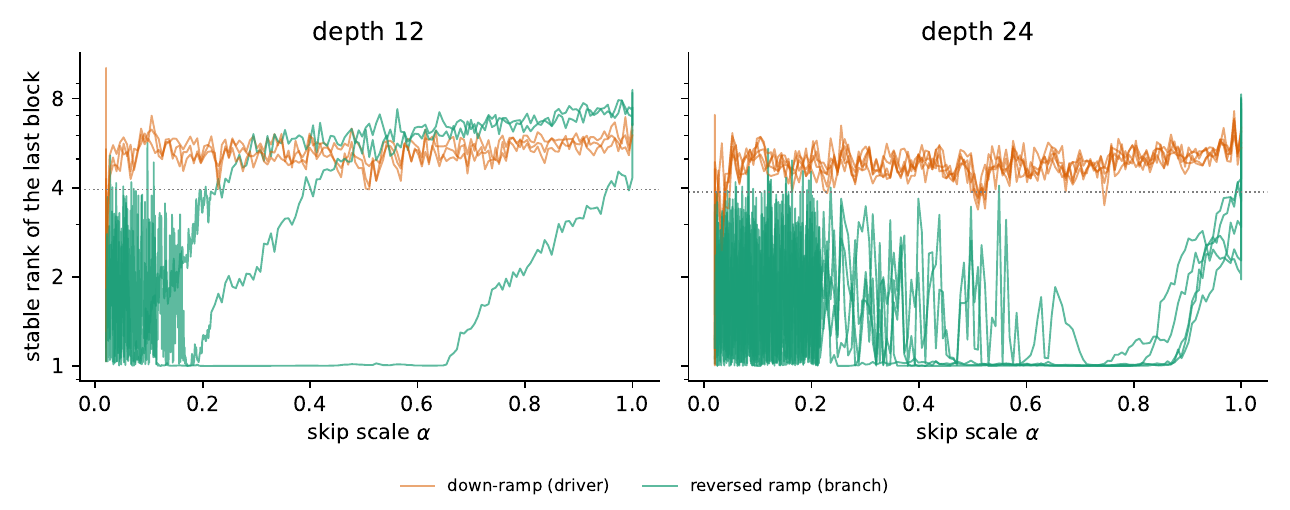}
  \caption{Stable rank of the last block against the skip scale $\alpha$ while it is ramped down (drivers) and,
  from the saved collapsed state, while it is ramped back up at the same rate (main study).  Dotted line:
  recovery threshold.}
  \label{fig:hysteresis}
\end{figure}

If the path is what fails, reopening it should help where added force did not.  Restoring the skip connection
does exactly that without touching a weight.  On the reversed ramp every network recovered its rank,
\RecRampup/\NRampup{} in the main study and \RepRampupRec/\RepRampupN{} in the replication.  Setting $\alpha$ back
to a fixed value at once, a \emph{switch}, reopened the path immediately (\cref{fig:path}, bottom): at the first
step after $\alpha$ was set to 1, before any update,
the task gradient on the query and key weights rose by a factor of about $10^{\PathSwitchOneRiseMed}$ even in
the block where it rose least, and in the median block it reached \PathSwitchOneVsHealthyMed{} times the
healthy value.  The rank itself had not yet recovered at that moment in most networks; the recovery was
learning made possible by the reopened path.

Recovery did not happen where the collapse had begun.  The collapse appeared when $\alpha$ reached
\AlphaDetMin--\AlphaDetMax, but on the way back the rank recovered only at $\alpha$ between
\AlphaRecTwelveMin{} and \AlphaRecTwelveMax{} at depth 12 and at $\alpha=\AlphaRecTwentyFourMin$ at depth 24
(\cref{fig:hysteresis}; replication \RepAlphaRecTwelveMin--\RepAlphaRecTwelveMax{} and
\RepAlphaRecTwentyFourMin), at least \HystMinRatio{} times above the point of collapse in all \HystEventsN{}
events.  This is a hysteresis.  Because the ramp also lets time pass, we checked it by setting $\alpha$ to a
fixed value at once: at 0.3 and 0.6, \HoldThreeRec{} of \HoldThreeN{} and \HoldSixRec{} of \HoldSixN{} networks
recovered within \HoldThreeTmin--\HoldThreeTmax{} steps, but at 0.1, five times the scale of collapse, only
\HoldOneRec{} of \HoldOneN{} did, and \PredPOneObsRec{} of \PredN{} in the replication.  On the way down every
network had still been healthy at that scale.  A restoration therefore has to go well beyond the point where
the collapse began.

Time mattered as well.  The same switch to 0.3 applied after 500, 1\,500 or 3\,000 further steps in the
collapsed state recovered later or not at all: at depth 12 the time to recovery rose with the delay in every
seed, and at depth 24 no delayed branch recovered in the main study or in the replication, where this was a
registered prediction (\cref{sec:posthoc}).  What changes with time was not identified; the attention pattern
did not.

Restoring the rank did not restore the loss.  All \SkipRecN{} networks that recovered by rank after the skip was
restored ended the budget \SkipRecGapMin--\SkipRecGapMax{} nats above the healthy network at the same step
(median \SkipRecGapMed).  Part of this gap may be the training time lost while collapsed.

\section{Does the path predict recovery?}
\label{sec:predict}

\begin{figure}[t]
  \centering
  \includegraphics[width=\linewidth]{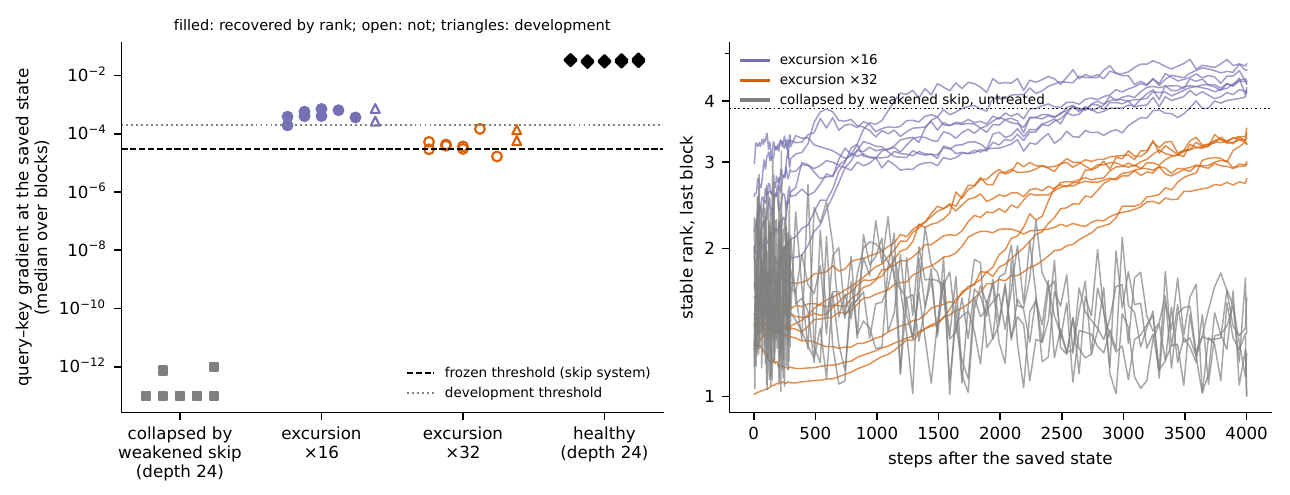}
  \caption{Registered test under a learning-rate excursion with the skip intact.  Left: query--key gradient at the
  saved state (median over blocks) for the depth-24 networks collapsed by a weakened skip (\cref{sec:where}), the held-out
  excursion networks at multipliers 16 and 32, and healthy depth-24 networks; filled: recovered by rank, open:
  not, triangles: development networks (\cref{app:registered}); dashed: threshold frozen in the first registered test; dotted: development
  threshold.  Right: stable rank of the last block after the saved state, excursion networks against the
  untreated depth-24 networks of the main study (50-step median; dotted: recovery threshold).}
  \label{fig:excursiontest}
\end{figure}

If the path decides, measuring it should tell in advance which networks will recover.  We tested this twice,
each time committing the predictions and the evaluation to the repository before the runs; the details are in
\cref{app:registered}.  The predictor was the median over blocks of the task gradient on the query and key
weights, read after the skip is restored and before any update.

On the ten replication networks the predictor ranked the recovery times after a switch to 0.3 as registered
(Spearman \PredPTwoRho{} over \PredPTwoN{} recovered networks).  Its threshold, fitted on the main study,
classified \PredPOneCorrect{} of \PredN{} outcomes at the weaker switch to 0.1, one short of the registered
target.  A simple state measure did as well: the loss gap, the excess of the collapsed network's loss over that of a
healthy one.

The second test changed the cause of the collapse.  On \ExcTestN{} new healthy depth-24 networks the learning
rate was multiplied by 16, or by 32, for 200 updates with the skip intact, and each network then continued
untreated.  The rank fell below the collapse threshold within ten updates, but the path stayed open: the query
and key gradient at the end of the burst was \ExcTestQkSixteenMin--\ExcTestQkSixteenMax{} at 16, far above the
networks collapsed by a weakened skip (\cref{fig:excursiontest}, left).  At 16 all \ExcTestRecSixteen{} networks
recovered their rank without treatment; at 32 none did within the budget, although their rank rose throughout
(\cref{fig:excursiontest}, right).  The networks collapsed by a weakened skip did not move in the same time.
The threshold from the first test did not carry over (\ExcTestTOneThirtyTwo{} of 8 correct at 32), and within
the networks at 16 the predictor did not order the recovery times (Spearman \ExcTestTThreeRho; the registered
main test was missed), whereas the loss gap did (\ExcTestTThreeLossRho).  Training stayed numerically stable,
and the networks that recovered their rank ended \ExcTestTFourGapMin--\ExcTestTFourGapMax{} nats above a copy
of the same network trained without the burst.

\section{Discussion}
\label{sec:discussion}

\paragraph{What decided repair.}  The added loss terms failed although their force was large, and they failed
because almost none of it reached the weights that had to change.  Restoring the skip connection changed no
weight but reopened the path, and the network then learned its way out.  When the collapse came from a burst of
high learning rate instead, the path stayed open and the rank recovered without help.  In attention-entropy
collapse a loss term of the bounded kind does repair \citep{hofmann2026repairing}; there the path was not
measured.  Together the cases separate the size of a corrective force from whether it arrives.  In rank
collapse arrival was the binding constraint.

\paragraph{What carries beyond this collapse.}  Any term added to a loss acts on the weights only through the
backward pass.  Its size at the loss says little about its effect if that pass no longer reaches the weights
the term is meant to move.  Measuring the gradient where it has to act, block by block, is cheap and needs no
training; here it distinguished the collapse driven by a weakened skip, in which the path was closed, from the
rank loss after a learning-rate burst, in which it stayed open.  It did
not, however, predict recovery better than a simple measure of the loss in our registered tests, so it is a
diagnostic of the mechanism rather than a better forecast.

\paragraph{A recovered diagnostic is not a recovered function.}  Every network in this paper that recovered by
rank still had a loss above its control at the end of the budget, after the skip was restored by
\SkipRecGapMin--\SkipRecGapMax{} nats and after the learning-rate burst by
\ExcTestTFourGapMin--\ExcTestTFourGapMax.  The rank is the measure on which collapse and recovery are defined
here, and it reported recovery while the function had not recovered.  Whether a run has recovered should be
judged by continuing it and measuring its loss.  For the learning-rate burst, a rollback to the checkpoint
before it would have cost 200 steps.

\section{Scope and limitations}
\label{sec:limitations}

This paper does not claim that loss-based repair fails for rank collapse in general.  The collapse of the main
study is driven by a weakened skip connection; with the skip intact, a weight-decay ramp and a larger
initialisation gain produced no collapse, and the learning-rate burst lowered the rank without closing the
path.  The networks have 9 to 11 million parameters and are trained on one data stream.  The paper does not
claim that the path measurement predicts recovery better than simple state measures, that a network that
recovered by rank has recovered its function, or what changes with time in the collapsed state.  The gradient
path was measured on one probe batch per network, the registered tests rest on ten and eight held-out networks
of one system, and the design was fixed before the main runs but not independently timestamped.  Alternative
explanations we considered are discussed in \cref{app:alternatives}.

\section{Conclusion}
\label{sec:conclusion}

A loss term can only repair what its gradient reaches.  In a depth-induced rank collapse the corrective force
of a well-designed term was large, but almost none of it reached the attention weights of the blocks that
complete the collapse.  Restoring the skip connection reopened that path and the network recovered its rank,
only well above the scale of collapse, without its loss returning to that of a healthy network.  Whether a
collapsed network came back depended on how much of the gradient's path was restored, not on what was added to
the loss.

\section*{Author contributions}

M.H.\ conceived the study, wrote the protocol and the code, ran and analysed the experiments and wrote the
manuscript.  P.M.\ supervised the work and revised the manuscript.

\section*{Reproducibility statement}
\label{sec:repro}

Code, preregistration, protocol, frozen constants, analysis scripts and the scripts that generate this
manuscript are kept in one version-controlled repository and are available from the authors on request.
Freeze commit \breakhash{\FreezeCommit}; preregistration SHA-256 \breakhash{\PreregSha}; calibration summary
SHA-256 \breakhash{\CalibrationSha}.  Every measured quantity in this paper is a macro rendered from the
analysis outputs; tables and figures are produced by the same scripts, and an audit checks hashes,
placeholders and unbound decimals before every build.  Training ran on NVIDIA A100-SXM4-40GB GPUs
(\GpuHours{} GPU-hours of wall time for calibration and main study at the cluster share available); the
gradient-path measurements ran on CPU from the saved states.  The data stream is the FineWeb-Edu sample
prepared with the GPT-2 tokenizer and a SHA-256 manifest by a sibling project.
The registrations of \cref{sec:predict} and \cref{app:registered} were committed before the runs they predict
(commits 6358641, b4b0963 and bf90737).

\section*{Use of AI-assisted tools}

Claude Code (Anthropic; Claude Fable 5.1 and Claude Opus 5.5, September and October 2026) assisted with
research code, run orchestration, analysis scripts, literature lookup and drafting.  The authors designed and
verified the study and are responsible for its interpretation, citations and final text.

\bibliography{references}
\bibliographystyle{plainnat}

\clearpage
\appendix
\crefalias{section}{appendix}

\section{Design, calibration and detection}
\label{app:design}

\begin{figure}[t]
  \centering
  \includegraphics[width=\linewidth]{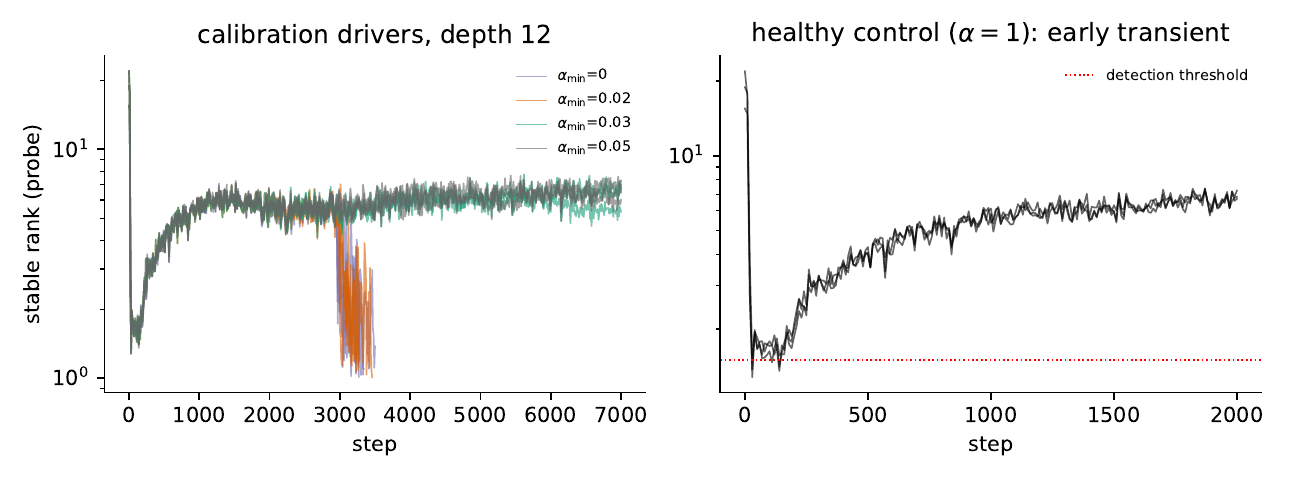}
  \caption{Left: depth-12 drivers over the $\alpha_{\min}$ grid; at this depth only $\alpha_{\min}\le\AlphaMin$
  collapses, and it does so when $\alpha$ reaches the floor.  Right: the healthy network ($\alpha=1$) dips to
  the detection threshold early in training and heals by itself.}
  \label{fig:calibration}
\end{figure}

\paragraph{Driver.}  The floor $\alpha_{\min}=\AlphaMin$ is the largest value on a calibration grid at which
every seed collapsed and no untreated network recovered.  At $\alpha_{\min}=0.03$ every depth-12 network kept
its rank, while at depth 24 some seeds lost it, one of them only briefly; above $0.03$ none did.  The driver is
a knife-edge (\cref{fig:calibration}).

\paragraph{Detection, settling and recovery.}  The collapse threshold is a stable rank below 1.5 on the last
block.  Because the collapse oscillates for a few hundred steps after it first appears, a network is declared
settled, and saved, at the first measurement at least 100 steps later whose ten-measurement median is still
below 1.5.  Recovery is declared when the 50-step median of the stable rank reaches half its healthy value
($\SrankRecTwelve$ at depth 12, $\SrankRecTwentyFour$ at depth 24).  Of the ten main-study collapse events,
\NumSettled{} settled and \NumTransient{} (both at depth 12) dipped below the threshold for one to three
measurements and healed by themselves; those two were never treated.  Healthy networks dip below the collapse
threshold early in training (\HealthyDipN{} of \HealthyN{} control runs, all within the first \HealthyDipMax{}
steps) and reach the recovery threshold by step \HealthyRecMax; the ramp starts late enough that this transient
is never mistaken for the collapse.

\paragraph{The collapsed state.}  The collapsed network is not simply the healthy network with a small skip.
The healthy weights evaluated at $\alpha=\AlphaMin$ give a loss of
\HealthyAtCollapsedLossTwelve{}--\HealthyAtCollapsedLossTwentyFour{} and a gradient norm near
\HealthyAtCollapsedGradTwelve, whereas the collapsed weights sit at the frequency-table loss with a gradient
norm of order one.

\paragraph{Loss terms.}  The hinge's gradient uses only the top singular pair, which is well defined as long as
$\sigma_1$ is separated from $\sigma_2$; since $\sigma_2\le\|E\|_F=\sigma_1\Q$, it is separated at the collapse.
The loss actually added on the first training step was within a tenth of its target in \MatchedWithinTen{}
loss-treated branches.  The cosine term's gradient is neither fading nor constant near the collapse.  During
calibration, none of \CalFunctionForksN{} checks with the hinge on the last block only, on every block, or at
ten times $\Lambda$ recovered.

Calibration used seeds 101--103.  At $\alpha_{\min}=0.03$ \CalCollapseAtThree{} runs crossed the collapse
threshold, one of which settled, and none above did; $0.02$ and $0$ collapsed \CalCollapseAtTwo{} and
\CalCollapseAtZero{}, and \CalNoneForks{} untreated branches stayed collapsed.  Changes made during calibration
and before the main runs: the ramp start moved from step 200 to \RampStart{} because of the healthy network's
early dip; settling and recovery use medians because the collapse oscillates; the loss terms act on every
block rather than the last one only, because a last-block term failed identically for both classes; the
$\alpha_{\min}$ grid was widened to locate the knife-edge; the cosine term was labelled class-less; the
interpretation row for ``both treatments fail'' was added.  Additions after the main runs are reporting only:
the settled/self-healing classification, the $\rho$ table, the per-block gradient measurements on all saved
collapsed networks and after the switch, and the post hoc branches.  The branches on the replication networks
in \cref{app:registered} were registered before they ran.

\section{Force measurement}
\label{app:force}

\begin{figure}[t]
  \centering
  \includegraphics[width=\linewidth]{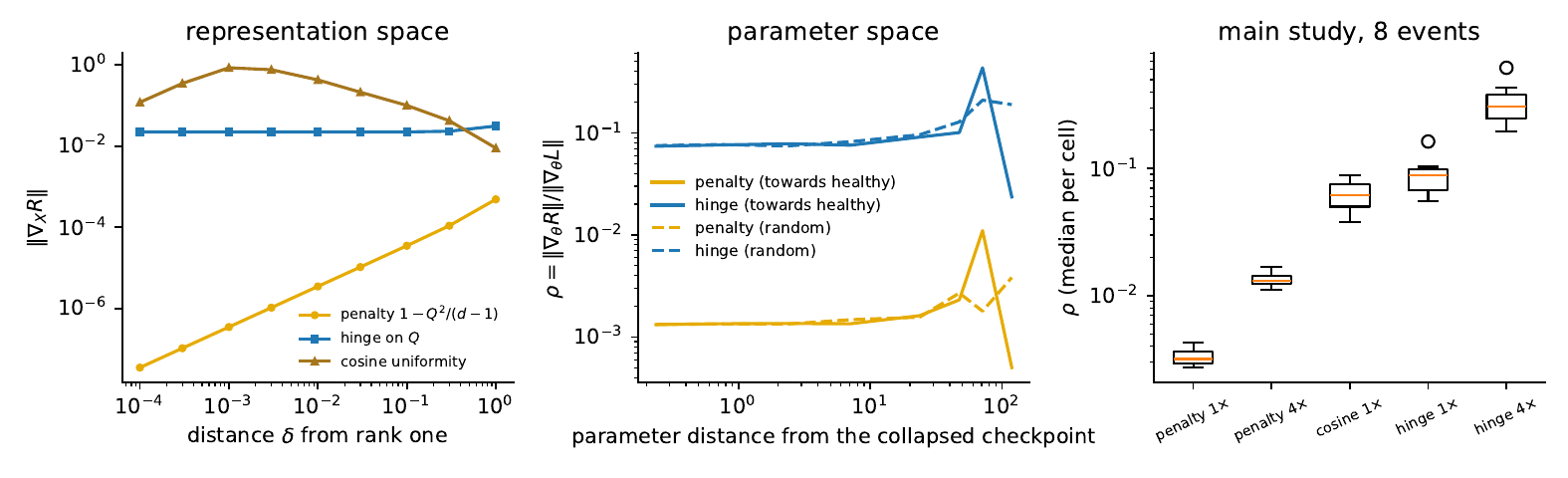}
  \caption{Force of each loss term.  Left: gradient norm with respect to the synthetic token matrix
  as its distance $\delta$ from rank one shrinks; the penalty fades linearly, the hinge stays constant, the
  cosine term is non-monotone because its row normalisation diverges for tokens near the mean.  Middle:
  ratio $\rho$ of added-loss to task gradient at the parameters of a collapsed calibration checkpoint and at
  points along the straight line towards the healthy network of the same seed and, as a control, the same distance along a
  random direction.  Right: $\rho$ per treatment over the eight treated collapse events.}
  \label{fig:force}
\end{figure}

On a synthetic matrix $X=av^\top+\delta E$, in which every token lies on one direction $v$ and a random matrix
$E$ scaled by $\delta$ sets the distance from rank one, the penalty's gradient falls from \AuditPenForceHigh{} at
$\delta=\AuditDeltaHigh$ to \AuditPenForceLow{} at $\delta=\AuditDeltaLow$, while the hinge's stays at
\AuditHingeForce{} (\cref{fig:force}, left).  In the audit that preceded the main runs the ratio $\rho$ at a
collapsed calibration checkpoint was \AuditRhoPen{} for the penalty and \AuditRhoHinge{} for the hinge.  The
cosine term sat between the two (\RhoCosOne).  \Cref{tab:rho} gives the ratio per treatment.

\begin{table}[t]
  \centering
  \caption{Ratio $\rho$ of added-loss to task gradient norm per treatment (median over measurements per
  collapse event; median and range over the eight events) and the fraction of steps on which gradient
  clipping was active.}
  \label{tab:rho}
  \begin{tabular}{lrrrr}
\toprule
Treatment & events & $\rho$ median & range over events & clip fraction \\
\midrule
penalty 1$\times$ & 8 & 0.0032 & 0.0027--0.0043 & 0.22 \\
hinge 1$\times$ & 8 & 0.0887 & 0.0550--0.1630 & 0.44 \\
penalty 4$\times$ & 8 & 0.0130 & 0.0111--0.0170 & 0.21 \\
hinge 4$\times$ & 8 & 0.3071 & 0.1942--0.6194 & 0.60 \\
cosine 1$\times$ & 8 & 0.0618 & 0.0380--0.0881 & 0.40 \\
\bottomrule
\end{tabular}

\end{table}

\IfFileExists{generated/posthoc_section.tex}{\section{Post hoc checks}
\label{sec:posthoc}

The following checks were run after the main analysis and are labelled post hoc.  They use the recovery rule fixed before the main runs but were not part of the design, and they are not pooled with the main result.  Branches run: immediate switch 24 of 24, delayed switch 24 of 24, unclipped added loss 32 of 32; intact-skip drivers 24 of 24.
\paragraph{Immediate switch.}
Branches of the eight saved collapsed networks with the skip scale set to a fixed value for the full budget:
\begin{center}\begin{tabular}{lrrl}
\toprule
$\alpha$ held & recovered / n & median $T_{\mathrm{rec}}$ (recovered) & final srank range \\
\midrule
0.1 & 2 / 8 & 94 & 1.00--7.09 \\
0.3 & 7 / 8 & 71 & 1.48--7.18 \\
0.6 & 8 / 8 & 50 & 6.40--7.42 \\
\bottomrule
\end{tabular}\end{center}
\paragraph{Delayed switch.}
Branches kept at $\alpha_{\min}$ for a further 500, 1\,500 or 3\,000 steps before the switch to $\alpha=0.3$ (the same switch as in the undelayed branches above):
\begin{center}\begin{tabular}{lrrl}
\toprule
steps at $\alpha_{\min}$ before the switch to 0.3 & recovered / n & median $T_{\mathrm{rec}}$ after the switch & final srank range \\
\midrule
500 & 2 / 8 & 150 & 1.04--6.86 \\
1500 & 3 / 8 & 290 & 1.00--6.95 \\
3000 & 1 / 8 & 3770 & 1.02--3.60 \\
\bottomrule
\end{tabular}\end{center}
\paragraph{Added-loss gradient exempt from clipping.}
With the task gradient clipped alone and the added-loss gradient added unclipped: penalty 1$\times$: 0/8 recovered (final srank 1.02--1.99); hinge 1$\times$: 0/8 recovered (final srank 1.00--2.60); hinge 4$\times$: 0/8 recovered (final srank 1.01--3.63); hinge 10$\times$: 0/8 recovered (final srank 1.41--3.40).
\paragraph{Intact-skip drivers (exploratory).}
Attempts to produce a collapse with the skip connection intact ($\alpha\equiv1$), with an uncentred stable rank as the detector, over 9000 steps each: weight-decay ramp to 10: detected 0/6, settled 0/6; initialisation gain 4: detected 0/6, settled 0/6; initialisation gain 8: detected 0/6, settled 0/6; initialisation gain 16: detected 0/6, settled 0/6.  Neither driver produced a settled collapse, so no branches were made from them.  
}{}

\begin{figure}[t]
  \centering
  \includegraphics[width=0.6\linewidth]{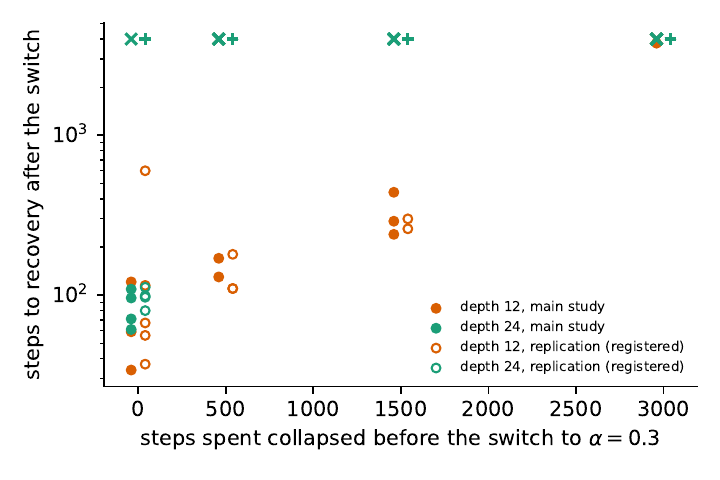}
  \caption{Filled: main study, post hoc.  Open: the same switches on the replication networks, registered
  before they ran.  Steps to recovery after the switch to $\alpha=0.3$ against the steps
  spent at $\alpha_{\min}$ before the switch (0: undelayed).  Crosses at the top: no recovery within the
  budget.}
  \label{fig:age}
\end{figure}

\paragraph{Recovery after delay.}  This section reports a post hoc observation of the main study.  The same switch to
$\alpha=0.3$, applied after the branch has spent a further 500, 1\,500 or 3\,000 steps at $\alpha_{\min}$,
took longer or failed (\cref{fig:age}; pooled medians and final stable ranks in the table above).  At
depth 12, \HoldThreeRecTwelve/\HoldThreeNTwelve{} recovered without delay,
\AgeFiveRecTwelve/\AgeFiveNTwelve{} after 500 steps (in \AgeFiveTmin--\AgeFiveTmax{} steps),
\AgeFifteenRecTwelve/\AgeFifteenNTwelve{} after 1\,500 (in \AgeFifteenTmin--\AgeFifteenTmax) and
\AgeThirtyRecTwelve/\AgeThirtyNTwelve{} after 3\,000 (at \AgeThirtyTmin).  Within each seed the recoveries
took longer the longer the delay.  At depth 24 no delayed branch recovered
(\AgeFiveRecTwentyFour/\AgeFiveNTwentyFour, \AgeFifteenRecTwentyFour/\AgeFifteenNTwentyFour,
\AgeThirtyRecTwentyFour/\AgeThirtyNTwentyFour), although \HoldThreeRecTwentyFour/\HoldThreeNTwentyFour{}
recovered without delay.  With so few events per delay and depth the counts alone are weak evidence.  The observation
rests on the depth-12 recovery times, which rose with the delay without overlap among the events that
recover, and on the depth-24 contrast between the undelayed switch and every delayed one.  A delayed full
restoration was not run.  The reversed ramp is partly a delayed case: by the time it has raised $\alpha$
back to 0.3 the network has spent about \RampUpStepsToThree{} further steps collapsed.

What changes with time is not the attention pattern: its normalised entropy was unchanged after 3\,500
untreated steps (\EntLateMin--\EntLateMax).

\IfFileExists{generated/replication_section.tex}{\section{Replication with new seeds}
\label{sec:replication}

The same design, rules and data order were run once more with five new seeds per depth.  The results are reported next to the main study, not pooled with it.  Of 10 collapse events, 10 settled and none healed by itself.  
Again no loss-based treatment recovered the rank within the budget, and neither did the untreated network (0 of 10 in each of the 6 treatments).  
At the parameters the penalty delivered a force of 0.0037 of the task gradient and the hinge 0.0782 (medians over collapse events).  
On the reversed ramp every network recovered, but only after $\alpha$ had risen far past the value at which it collapsed: at $\alpha$ between 0.95 and 1.00 at depth 12 (5 of 5) and at $\alpha=1.00$ at depth 24 (5 of 5).  
The outcome fixed in advance for the two loss terms, both failing while the untreated network stays collapsed, was reproduced.
}{}

\section{Registered tests in detail}
\label{app:registered}

\paragraph{Path after the switch.}  The rule fixed for this check before it was run, a query--key gradient of at least
$10^{-2}$ in every block, was met at $\alpha=1$ in \PathSwitchOneStepZeroOk{} of \PathN{} networks and at
$\alpha=0.3$ in \PathSwitchThreeStepZeroOk.  The healthy networks themselves met it in \PathHealthyQkMinOk{}
of \PathN, so the rule was stricter than the healthy state.  On the attention input weights, the quantity
of the profile in \cref{sec:where}, it was met in \PathSwitchOneStepZeroAttnOk{} of \PathN{} networks at
$\alpha=1$ and in \PathSwitchThreeStepZeroAttnOk{} at $\alpha=0.3$.  The switch by itself restored the rank only in part:
before any update the stable rank of the last block lay above the recovery threshold in
\PathSwitchThreeStepZeroAboveRec{} of \PathN{} networks at $\alpha=0.3$ and in
\PathSwitchOneStepZeroAboveRec{} at $\alpha=1$.  The rest of the recovery was learning made possible by the
reopened path.  After 50 updates the rank lay above the recovery threshold in
\PathSwitchThreeAfterFiftyAboveRec{} of \PathN{} networks at $\alpha=0.3$ and in
\PathSwitchOneAfterFiftyAboveRec{} at $\alpha=1$.  These updates were replayed from the saved
weights, optimizer state and data position.  Where the logged main-study branches were measured at the same
steps, the replayed stable rank agreed with the logged one to \PathForkCheckMedPct\,\% in the median.

\paragraph{First test: the replication networks.}  The gradient path of \cref{sec:runtime} is read before any update, so it can predict what a restoration will
do.  We tested this on the ten saved networks of the replication, on which no switch had been run.  Before
any branch ran, the predictor, four predictions and the evaluation script were committed to the repository.
The predictor is the median over blocks of the task gradient on the query and key weights at the first step
after the switch, before any update.  For the switch to $\alpha=0.1$ its threshold was fitted on the eight
main-study networks, which it separates without error (\PredMainAcc), and frozen at \PredThreshold.

The predictions were these.  P1: at $\alpha=0.1$ exactly the networks above the threshold recover; met if at
least 9 of 10 are classified correctly.  P2: at $\alpha=0.3$ the predictor ranks the recovery times, with a
Spearman correlation of $-0.4$ or lower (\PredDevRho{} on the main study).  P3: the force of the hinge, its
ratio $\rho$, does not predict these outcomes.  P4: rival measures of the state predict worse: the loss gap
at detection, the task gradient norm at the saved state, and depth alone.  Depth matters because the
predictor is larger at depth 12; the predictor and depth disagree on two of the ten networks.

\begin{figure}[t]
  \centering
  \includegraphics[width=\linewidth]{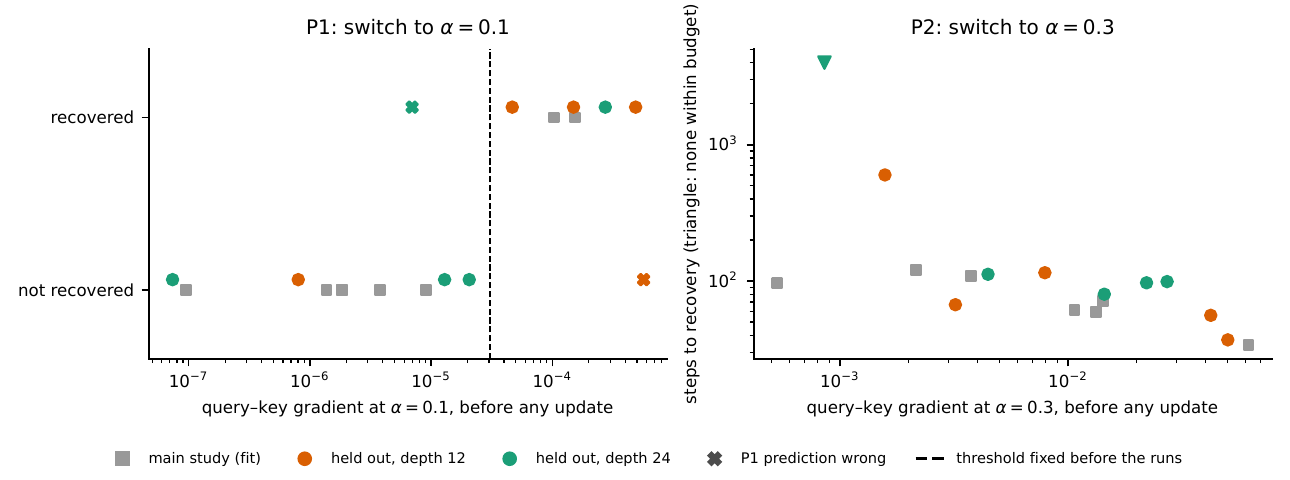}
  \caption{Registered prediction on the ten replication networks.  Left: outcome of the switch to
  $\alpha=0.1$ against the predictor; dashed line: threshold fitted on the main study (grey squares) and frozen
  before the branches ran; crosses: wrong predictions.  Right: steps to recovery after the switch to
  $\alpha=0.3$ against the predictor at that scale.}
  \label{fig:prediction}
\end{figure}

P1 was missed by one network: \PredPOneCorrect{} of \PredN{} were classified correctly (\cref{fig:prediction},
left).  At $\alpha=0.1$, \PredPOneObsRecTwelve{} of \PredNTwelve{} networks recovered at depth 12 and
\PredPOneObsRecTwentyFour{} of \PredNTwentyFour{} at depth 24.  So, as in the main study (\cref{sec:runtime}),
half or more of the networks stayed collapsed at five times the scale of collapse without any delay.  One error is the depth-12 network whose
query--key gradient was not near zero in the collapsed state (\cref{sec:where}); it lay above the threshold
and did not recover.  The other is a depth-24 network below the threshold that recovered.  On the two networks
where the predictor and depth disagree, the predictor was right both times; depth alone classified
\PredRivalDepthAcc{} of 10.  P2 was met: \PredPTwoN{} networks recovered at $\alpha=0.3$, and the Spearman
correlation between predictor and recovery time was \PredPTwoRho{} (\cref{fig:prediction}, right).  P3 was
missed.  The hinge force classified \PredRivalHingeAcc{} of 10 at $\alpha=0.1$, but it correlated with the
recovery time at \PredRivalHingeRho, above the registered bound of 0.4 in magnitude: a larger force went with
slower recovery.  P4 was missed.  The gradient norm did worse (\PredRivalGradNormAcc{} of 10, Spearman
\PredRivalGradNormRho), but the loss gap at detection did as well as the path predictor
(\PredRivalLossGapAcc{} of 10, Spearman \PredRivalLossGapRho).

The delayed switch of \cref{sec:runtime} was registered for the same networks.  The prediction was no recovery
at depth 24 at any delay, and at depth 12 recovery times that do not fall and recoveries that do not return as
the delay grows.  Both held (\cref{fig:age}).  At depth 24 none of the 15 delayed branches recovered, although
\RepHoldThreeRecTwentyFour{} of 5 recovered without delay.  At depth 12, \RepHoldThreeRecTwelve{} of 5
recovered without delay, \RepAgeFiveRecTwelve{} after 500 steps (in \RepAgeFiveTmin--\RepAgeFiveTmax{} steps),
\RepAgeFifteenRecTwelve{} after 1\,500 (in \RepAgeFifteenTmin--\RepAgeFifteenTmax) and
\RepAgeThirtyRecTwelve{} after 3\,000, and within each seed the time to recovery rose with the delay.

\paragraph{Second test: a learning-rate burst.}  The predictor of the first test and its threshold were fixed on networks whose collapse was driven by a
weakened skip.  We tested whether both transfer unchanged to a different driver.  On \ExcTestN{} healthy
depth-24 networks that had not been used before, the learning rate was multiplied by 16, and in a second
condition by 32, for 200 updates with the skip intact.  Each network was saved at the end of the excursion and continued
untreated for \Budget{} steps; a paired branch at the base rate was the control.  The query--key gradient was
measured at the saved state, and the prediction for each network was committed before its continuation ran.
Two further networks, seeds 2 and 4, served for development.  They set the second condition, development thresholds for the path
and for two state measures, and the targets.  The two state measures, taken at the saved state, are the loss
gap, here the probe loss above that of the paired control, and the norm of the task gradient.  The registration fixed seven tests.  T0 asks whether the path is
open, T1 whether the frozen threshold classifies recovery, T1b and T2 whether the development threshold does
and whether it beats the state measures and the multiplier alone.  T3, the main test, asks whether the path
orders the recovery times within one multiplier, and T3r whether the state measures order them worse.  T4 asks
whether networks that recover by rank are still harmed, that is, end the budget with a loss on unseen data more
than 0.1 nats above the paired control.

The path stayed open.  At the saved state the query--key gradient was \ExcTestQkSixteenMin--\ExcTestQkSixteenMax{}
at multiplier 16 and \ExcTestQkThirtyTwoMin--\ExcTestQkThirtyTwoMax{} at 32, far above the networks collapsed by
a weakened skip and below healthy ones (\cref{fig:excursiontest}, left).  All \ExcTestTZero{} networks at 16 lay
above the frozen threshold (T0, target 7 of 8, \ExcTestTZeroMet).  At 16 all \ExcTestRecSixteen{} recovered by
rank, after \ExcTestTrecSixteenMin--\ExcTestTrecSixteenMax{} steps.  At 32 none did within the budget, although
the rank rose throughout and ended at \ExcTestEndThirtyTwoMin--\ExcTestEndThirtyTwoMax{}
(\cref{fig:excursiontest}, right).  Training stayed numerically stable at both multipliers; the harm was functional,
not a divergence.  The networks collapsed by a weakened skip did not move in the same budget.

The threshold did not transfer.  It classified \ExcTestTOneSixteen{} of 8 networks correctly at 16 and
\ExcTestTOneThirtyTwo{} of 8 at 32 (T1, target 7 per condition, \ExcTestTOneSixteenMet{} and
\ExcTestTOneThirtyTwoMet{} as registered).  The development threshold classified \ExcTestTOneB{} of
\ExcTestTOneBN{} (T1b, target 13, \ExcTestTOneBMet).  But so did the multiplier alone, and the development
thresholds of the loss gap and the gradient norm classified \ExcTestTTwoLoss{} and \ExcTestTTwoGrad{}.  A strict
win over every rival was therefore \ExcTestTTwoMet{} (T2).  Within the condition at 16, where the time to recovery varied, the
query--key gradient did not order it: the Spearman correlation was \ExcTestTThreeRho{} over \ExcTestTThreeN{}
recovered networks (T3, the main test, target $-0.4$ or lower, \ExcTestTThreeMet).  The loss gap ordered it
better (\ExcTestTThreeLossRho; a larger gap went with slower recovery), and so did the gradient norm (\ExcTestTThreeGradRho; T3r, \ExcTestTThreeRMet).
Over both conditions together every measure ranked the outcome (query--key gradient \ExcTestTThreePQk, loss gap
\ExcTestTThreePLoss, multiplier \ExcTestTThreePMult), because each separates the two conditions.

The rank recovered, the loss did not.  All \ExcTestTFourRec{} networks that recovered by rank ended the budget
\ExcTestTFourGapMin--\ExcTestTFourGapMax{} nats above the paired control (T4, target three quarters,
\ExcTestTFourMet).  Under this driver the path stayed open and the rank recovered untreated, as the path account
predicts.  The threshold fitted on the skip system did not carry over, and the path measurement did not order
recovery better than the loss gap.  The excursion left a loss that the rank does not show.

\section{Alternative explanations}
\label{app:alternatives}

\begin{itemize}
\item That the loss-based treatments fail for a reason of protocol rather than mechanism.  The unclipped
  branches of \cref{sec:repair} answer this.
\item That the hysteresis is an effect of time spent collapsed rather than a shifted skip scale of recovery.
  The immediate switches answer this: at 0.3 and 0.6 most networks recover within about a hundred steps, and
  at 0.1 fewer than half recover although no time has passed.
\item That the force measurement is true by construction.  In representation space it is.  The ratio at the
  parameters is not, and a single collapse event with a large penalty ratio would have overturned the
  classification.
\item That the two self-healing dips reflect a detector tuned to the data.  The replication under the same
  rules had none.
\item That a collapse driven by removing the skip is trivial.  This is true of its existence
  \citep{dong2021attention}, not of its persistence or of the failure of loss-based repair.
\item That the result holds only because the skip was weakened.  Under a learning-rate excursion with the skip
  intact the path stayed open and the rank recovered untreated, as the path account predicts
  (\cref{sec:predict}).  A settled collapse with the skip intact was not produced, so whether such a
  collapse would close the path is untested.
\end{itemize}

\section{Interpretation fixed in advance}
\label{app:claims}

Before the main runs, the outcome ``both loss-based treatments fail while
the untreated network stays collapsed and the loss stays flat'' was mapped to the reading ``the force
classification carries over, the repair does not'', and it is the outcome of the frozen comparison.  A
secondary prediction concerned three markers of the collapse along the ramp, the spectral gap
$\sigma_2/\sigma_1$, the mean squared cosine $U$ and the loss gap to the healthy network, each timed at the
first measurement where it is half-way from its value before the ramp to its value at detection.  The
predicted order, gap before $U$ before loss gap, occurred in \MarkerOrderOk{} events, and no order is
resolvable at this ramp rate.  The crossings fall at steps \MarkerGapMin--\MarkerGapMax,
\MarkerUMin--\MarkerUMax{} and \MarkerDlossMin--\MarkerDlossMax, all within a few measurements of the moment
$\alpha$ reaches its floor, with the loss gap no later than either spectral marker in \LossNotLater{} events.

\end{document}